\documentclass[letterpaper, 10 pt, conference]{ieeeconf}  

\IEEEoverridecommandlockouts                              
\title{\LARGE \bf
Delay-Aware Distributed Multi-Robot Motion Planning}
\author{Anonymous}
 \author{Ardalan Tajbakhsh$^{1}$, Augustinos D. Saravanos$^{2}$, James Zhu$^{3}$, \\
 Evangelos A. Theodorou$^{2}$, Lorenz T. Biegler$^{4}$, Aaron M. Johnson$^1$
\thanks{This work used Delta CPU at NCSA through allocation CIS240861 from Advanced Cyberinfrastructure Coordination Ecosystem: Services \& Support (ACCESS), which is supported by U.S. National Science Foundation grants \#2138259, \#2138286, \#2138307, \#2137603, and \#2138296.}%
\thanks{$^1$ Department of Mechanical Engineering and $^4$ Department of Chemical Engineering, Carnegie Mellon University, Pittsburgh, PA, USA, \texttt{\{atajbakh, lb01, amj1\}@andrew.cmu.edu}}%
\thanks{$^2$ Daniel Guggenheim School of Aerospace Engineering, Georgia Institute of Technology, Atlanta, GA, USA, \texttt{\{asaravanos, evangelos.theodorou\}@gatech.edu}}
\thanks{$^3$ Department of Electrical and Computer Engineering, University of Miami, Coral Gables, FL, USA, \texttt{jxz1636@miami.edu}}
}
\usepackage{graphicx}
\usepackage{subcaption}
\usepackage{amsmath}
\usepackage{mathtools}
\usepackage{amsthm}
\usepackage{amssymb}
\usepackage{accents}
\usepackage{cite}
\usepackage{float}
\usepackage{booktabs}
\usepackage{romannum}
\usepackage{amssymb} 
\usepackage{pifont}
\newcommand{\xmark}{\textcolor{red}{\ding{55}}}             

\usepackage{amssymb}    
\usepackage{pifont}     
\usepackage{tabularx}   
\usepackage{booktabs}   
\usepackage[dvipsnames]{xcolor}     
\usepackage{multirow}
\usepackage{cancel}

\newcommand{\pgain}[1]{\textcolor{green!60!black}{(#1)}}
\newcommand{\ploss}[1]{\textcolor{red!70!black}{(#1)}}

\newcommand{\cmark}{\textcolor{green!70!black}{\checkmark}} 
\DeclareMathOperator{\argmin}{argmin}
\usepackage{algorithmic}
\usepackage[ruled, lined, linesnumbered, longend]{algorithm2e}

\usepackage{hyperref}
\hypersetup{colorlinks,allcolors=black}
\SetKwInput{KwInput}{Input}                
\SetKwInput{KwOutput}{Output}              
\hypersetup{%
  breaklinks,
  bookmarksnumbered=true,
  bookmarksopen=true,
  colorlinks=true,
  linkcolor=black,
  urlcolor=black,
  citecolor=black
}

\begin{document}
\bstctlcite{IEEEexample:BSTcontrol} 

\maketitle
\thispagestyle{empty}
\pagestyle{empty}

\begin{abstract}
This paper examines why distributed optimization degrades when multi-robot systems
coordinate under realistic communication delays. Using consensus ADMM as a
representative framework for generating collision-free, dynamically feasible motion
plans in both trajectory optimization and receding-horizon control settings, we show
that delay degrades performance structurally: the consensus and dual updates weight
stale and current information identically, so staleness enters the optimizer disguised
as genuine disagreement and accumulates through the multipliers. Delay-agnostic
adaptation (e.g. residual balancing) inherits this contamination and can perform worse
than a fixed penalty. We introduce a Delay-Aware ADMM (DA-ADMM) variant that adapts
penalty parameters to real-time delay statistics and re-derives the global consensus
update as a delay-weighted average, down-weighting stale information in favor of recent
updates. Simulations in 2D and 3D show that DA-ADMM improves robustness, success rate,
and solution quality over fixed-parameter, residual-balancing, lower-bound, and
fixed-constraint baselines. Degradation is therefore governed not by delay length or
frequency, but by the optimizer's ability to reason over the age of the information it
receives.
\end{abstract}
\section{Introduction}
As robotic systems scale in dynamic environments such as warehouses, ports, and
sidewalks, distributed coordination under realistic deployment conditions becomes
critical. Such settings demand algorithms that reason over realistic agent dynamics,
handle tight inter-agent conflicts, and remain robust to imperfect communication
\cite{gielis2022critical}.

Much of the existing work in multi-agent path finding (MAPF) relies on restrictive
assumptions on the agent dynamics and centralized, perfect communication
\cite{sharon2015conflict,barer2014suboptimal,boyarski2015icbs}. These approaches scale
to thousands of agents \cite{liu2020prediction} but often transfer poorly to the real
world. Extensions that plan in continuous time using model predictive control (MPC)
\cite{tajbakhsh2024conflict} or kino-dynamic RRT \cite{kottinger2022conflict} as the
low-level planner still require centralized coordination and synchronous communication.

Distributed optimization offers an attractive alternative at scale. Existing approaches
plan each agent's motion in parallel, with some treating the predictions of others as fixed
constraints \cite{luis2020online, tordesillas2021mader, firoozi2020distributed}. While
effective in sparse scenarios, such ``fixed constraint'' schemes become infeasible or
deadlock in constrained spaces, fail to reach fair compromise, and lack resilience to
delay due to their single-shot nature.
The Alternating Direction Method of Multipliers (ADMM) \cite{boyd2011distributed}
instead decomposes the global problem into local subproblems solved in parallel,
enforcing consensus via iterative multiplier updates over the coupling constraints.
Recent work leverages this framework for large-scale multi-robot motion planning in
deterministic settings \cite{saravanos2023distributed_ddp, shorinwa2024distributed} and
under dynamics uncertainty \cite{park2019distributed, saravanos2023hierarchical,
saravanos2024dmpcs}.

Nevertheless, these works generally assume ideal communication. \cite{ferranti2022distributed}
proposes an asynchronous ADMM variant using shifted predictions of delayed agents with
more conservative collision constraints, while \cite{guo2017impact} performs
global updates using only the neighbors that have responded, yielding slower convergence
and degraded solution quality from partial neighborhood information.
To improve convergence in general, parameter adaptation schemes such as residual balancing
\cite{wohlberg2017admm,xu2017adaptive, xu2017admm, mccann2024robust} have been
introduced. These perform well in synchronous regimes but are myopic to asynchronicity:
adjusting parameters solely from instantaneous residuals ignores message age and
conflates disagreement from stale updates with true consensus error. Data-driven schemes
learn tunable parameters \cite{noah2024distributed} or adaptation policies
\cite{saravanos2025deep} with substantial robustness and scalability, but applying them
to multi-robot coordination requires contextually rich data at scale, and training
cascaded networks remains challenging.

Despite prior work noting the importance of reasoning about imperfect communication
\cite{gielis2022critical,shorinwa2024distributed}, its impact on multi-robot
coordination remains underexplored. Two questions remain unresolved. First, why and
through what mechanism do communication delay patterns degrade distributed optimization
for multi-robot coordination? Second, what minimal change to the algorithmic structure
restores coordination \textit{performance} and \textit{robustness} under imperfect
communication?

 \begin{figure*}[t]
    \centering
    \includegraphics[width=.93\linewidth]{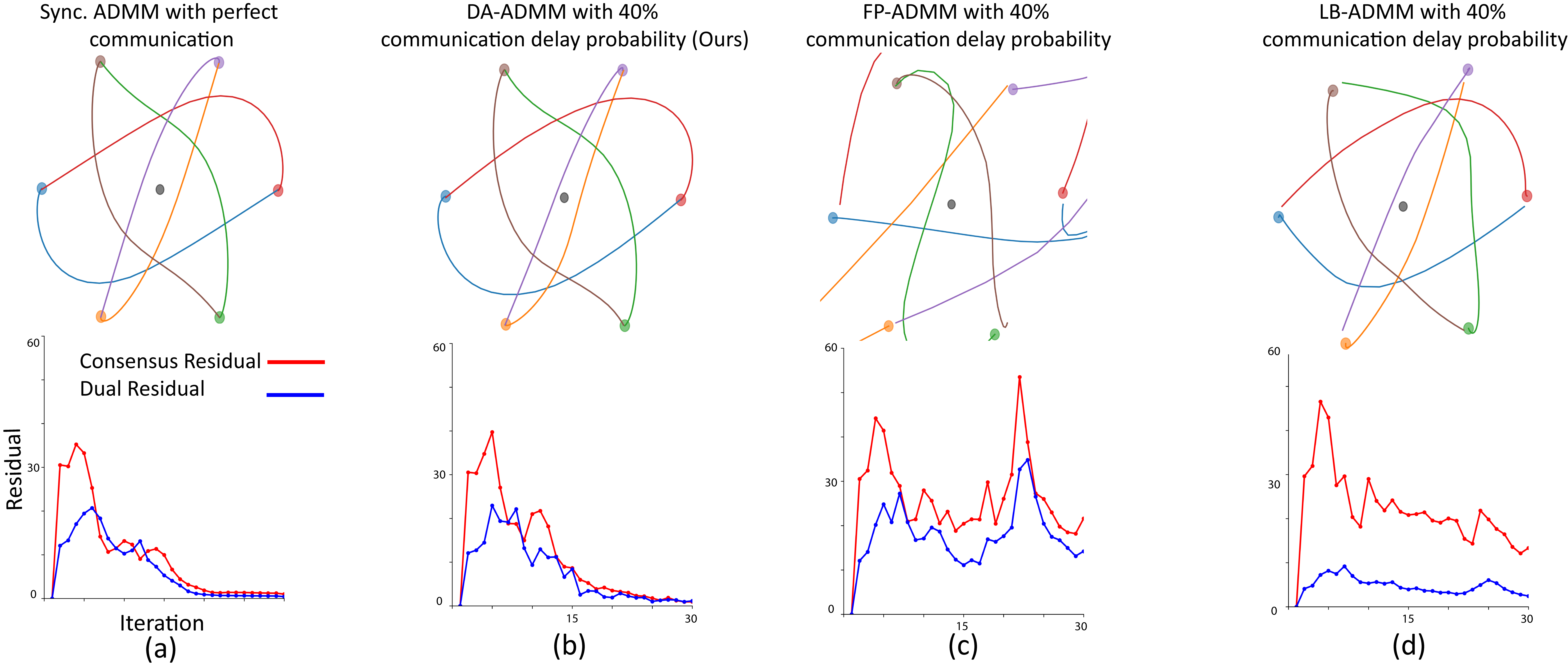}%
     \vspace{-.25em}%
    \caption{Comparison between different approaches for the 2D \textbf{Double Integrator Experiment}. Note that the proposed DA-ADMM with delayed communication (b) converges to a similar solution compared to the synchronous ADMM with perfect communication (a). Other fixed parameter ADMM baselines fail to converge under delayed communication (c and d).}
    \label{fig:experiment_1_trajopt}
    \vspace{-1em}
\end{figure*}

To address these questions, this paper makes the following contributions:
\begin{itemize}
\item We identify why bounded delay degrades consensus ADMM: uniform-weight averaging,
residual contamination, and dual accumulation. This is not a tuning artifact, and
delay-agnostic adaptation worsens it.
\item We quantify the effect of diverse delay patterns on success rate, convergence, and
solution quality, showing that degradation is governed not by delay length or frequency
but by whether the optimizer can reason over the age of its information.
\item We propose DA-ADMM, which corrects for this via delay-scaled penalties and a
re-derived weighted consensus update, recovering near-nominal performance.
\item We validate both the analysis and the correction in distributed trajectory
optimization and MPC settings across 2D and 3D scenarios (Fig.~\ref{fig:experiment_1_trajopt}).
\end{itemize}
\section{Problem Description}

\subsection{Problem Setup}
This section summarizes the problem formulation and notation, summarized in Table \ref{tab:notation_summary}. Consider a team of $N$ agents indexed by $\mathcal{I} := \{1, \dots, N\}$. Each agent $i \in \mathcal{I}$ has a set of neighbors $\mathcal{N}_i$ including itself. Each agent has a state $x_{i,k}$ and a control input $u_{i,k}$ at time step $k \in \mathcal{K} := \{0, \dots, K\}$. We define the state trajectory for agent $i$ as
$x_i := [x_{i,k}]_{k \in \mathcal{K}}$
and similarly the control trajectory as $u_i := [u_{i,k}]_{k \in \mathcal{K}}$.

Consider the following multi-agent optimization problem:
\begin{subequations}\label{eq:initial_multi_agent_problem}
\begin{align}
\min_{\{x_i, u_i\}_{i \in \mathcal{I}}} \quad & \sum_{i \in \mathcal{I}} J_i(x_i, u_i) \\
\text{s.t.} \quad 
& b_{i,k}(u_{i,k}) \leq 0 && \text{(Actuation limits)} \label{eq:actuation_constraint} \\
& h_{i,k}(x_{i,k}) \leq 0 && \text{(Obstacle avoidance)} \label{eq:state_constraint} \\
& \begin{aligned}
   g_{ij}(x_{i,k}, x_{j,k}) \leq 0 \\
   \forall j \in \mathcal{N} \backslash \{i\}
  \end{aligned}
&& \text{(Inter-agent const.)} \label{eq:coupling_constraint}
\end{align}
\end{subequations}

Each agent's cost function is given by:
\begin{equation}
J_i(x_i,u_i) = \sum_{k = 0}^{K-1} \ell_i(x_{i,k}, u_{i,k}) + \phi_i(x_{i,K})
\end{equation}
where $\ell_i$ and $\phi_i$ denote the running cost and terminal cost of each agent, respectively.

A commonly used approach is to solve \eqref{eq:initial_multi_agent_problem} in parallel is treating neighbor states in \eqref{eq:coupling_constraint} as fixed predictions \cite{luis2020online}. While effective in sparse scenarios, this method suffers from deadlocks and infeasibility in dense settings due to lack of coordination. Moreover, solution quality can significantly vary across agents, and there is no mechanism for mutual compromise. This approach is also sensitive to communication delays since it relies on one-shot trajectory generation. 

\subsection{Distributed Consensus ADMM}

A more coordinated strategy is to reformulate the problem using consensus ADMM. Each agent $i$ maintains local copies of its own and neighbors' trajectories, denoted as
$\tilde{x}_{i,k} := [x_{j,k}^{i}]_{j \in \mathcal{N}_i}$ for the state trajectory of agent $j$ and 
$\tilde{u}_{i,k} := [u_{j,k}^{i}]_{j \in \mathcal{N}_i}$ for the control trajectory.
Global consensus variables $z_{j,k}$ and $w_{j,k}$ represent shared state and control variables. Each agent $i$ also constructs its copies, 
$    \tilde{z}_{i,k} := [z_{j,k}]_{j \in \mathcal{N}_i}$ and
$\tilde{w}_{i,k} := [w_{j,k}]_{j \in \mathcal{N}_i}$.

The consensus-based reformulation of \eqref{eq:initial_multi_agent_problem} is:
\begin{subequations}\label{eq:consensus_multi_agent_problem}
\begin{align}
\min_{\{x_i, u_i\}_{i \in \mathcal{I}}} \quad & \sum_{i \in \mathcal{I}} J_i(x_i, u_i) \\
\text{s.t.} \quad 
& \eqref{eq:actuation_constraint},~\eqref{eq:state_constraint},~g_i(\tilde{x}_{i,k}) \leq 0, & \forall i, k \in \mathcal{K} \\
& \tilde{x}_{i,k} = \tilde{z}_{i,k}, & \forall i, k \in \mathcal{K} \\
& \tilde{u}_{i,k} = \tilde{w}_{i,k}, & \forall i, k \in \mathcal{K}
\end{align}
\end{subequations}

Each ADMM \cite{boyd2011distributed} iteration performs the following steps:

\begin{enumerate}
    \item \textbf{Local update:} Each agent $i$ solves:
    \begin{subequations}\label{eq:local_augmented_lagrangian}
    \begin{align}
    \!\!\!\!\argmin_{\tilde{x}_i, \tilde{u}_i} \quad 
    & J_i(x_i, u_i) 
    + \frac{\rho}{2} \sum_{k \in \mathcal{K}} 
    \left\| \tilde{x}_{i,k} - \tilde{z}_{i,k} 
    + \frac{y_{i,k}}{\rho} \right\|^2 \nonumber \\
    & + \frac{\mu}{2} \sum_{k \in \mathcal{K}} 
    \left\| \tilde{u}_{i,k} - \tilde{w}_{i,k} 
    + \frac{\lambda_{i,k}}{\mu} \right\|^2 \\
    \text{s.t.} \quad 
    & \eqref{eq:actuation_constraint},~\eqref{eq:state_constraint},~g_i(\tilde{x}_{i,k}) \leq 0
    \end{align}
    \end{subequations}
    where $y_{i,k}$ and $\lambda_{i,k}$ are the dual variables and $g_i(\tilde{x}_{i,k})$ denotes the collision avoidance constraint on the augmented state of agent $i$, which includes its neighbors. 

   \item \textbf{Global update:} For each agent $j$ and time step $k$, the global consensus variables are computed by averaging the local copies of the state and control variables maintained by neighboring agents:
    \begin{equation}
    z_{j,k} \leftarrow \frac{1}{|\mathcal{N}_j|} \sum_{i \in \mathcal{N}_j} x_{j,k}^{i}, \quad
    w_{j,k} \leftarrow \frac{1}{|\mathcal{N}_j|} \sum_{i \in \mathcal{N}_j} u_{j,k}^{i}
    \end{equation}
    Here, $x_{j,k}^{i}$ and $u_{j,k}^{i}$ denote agent $i$'s local estimates of agent $j$’s state and control at time $k$.
    
    \item \textbf{Dual update:}

\end{enumerate}
    \begin{align}
    y_{j,k}^{i,(t+1)} \leftarrow y_{j,k}^{i, (t)} + \rho \left( x_{j,k}^{i,(t+1)} - z_{j,k}^{(t+1)} \right) \\
    \lambda_{j,k}^{i,(t+1)} \leftarrow \lambda_{j,k}^{i (t)} + \mu\left( u_{j,k}^{i,(t+1)} - w_{j,k}^{(t+1)} \right)
    \end{align}
The dual variables measure the consensus constraint violation levels, which regulates the subsequent local updates based on the magnitude of the penalty variables. Convergence is then monitored via the primal residual:
\begin{equation}
r_{i,k}^{(t)} := (\tilde{x}_{i,k}^{(t)} - \tilde{z}_{i,k}^{(t)}) + (\tilde{u}_{i,k}^{(t)} - \tilde{w}_{i,k}^{(t)}) \quad \forall i, k
\end{equation}
which is checked for convergence via $\|r^{(t)}\|_2 \leq \epsilon_{\text{pri}}$.

\section{Delay in Consensus ADMM}
\subsection{Communication Model}
Communication delays arise from network latency and asynchronous execution. Each ADMM
iteration involves two communication rounds: \textit{local-to-global} (LG), where agents
transmit updated local variables for the consensus update, and \textit{global-to-local}
(GL), where global variables are returned for the dual update. Let
$d_{ij}^{\text{LG},(t)}$ and $d_{ij}^{\text{GL},(t)}$ denote the staleness of the most
recent value agent $j$ holds from agent $i$ in the respective round at iteration $t$.
Rather than delaying each message independently, we model disruptions as bursts on each
directed link: a current link initiates an episode with probability $p$ and duration
$D_{ij} \sim \mathcal{U}\{1,\dots,d_{\max}\}$, during which the receiver reuses the last
value received and staleness accumulates,
\begin{equation}
    d_{ij}^{(t+1)} = d_{ij}^{(t)} + 1 \;\; \text{if} \;\; d_{ij}^{(t)} < D_{ij},
    \quad\text{else } 0
\end{equation}
Episodes are drawn independently across links and rounds, and a disrupted link contributes the same misaligned copy
for $D_{ij}$ consecutive iterations. Agent $i$ therefore builds its consensus vectors
from the most recently received global variables,
\begin{equation}
  \!\!\!\!\tilde{z}_{i,k}^{(t)} := \left[ z_{j,k}^{(t - d_{ji}^{\text{GL},(t)})}
  \right]_{j \in \mathcal{N}_i}\!, \quad
  \tilde{w}_{i,k}^{(t)} := \left[ w_{j,k}^{(t - d_{ji}^{\text{GL},(t)})}
  \right]_{j \in \mathcal{N}_i}  \!\!\!\!\!\!\!
\end{equation}
while the global update for agent $j$ uses the local copies it has received:
\begin{align}
    z_{j,k}^{(t+1)} \!\!\!\leftarrow \!\!\frac{1}{|\mathcal{P}_j|} \!\sum_{i \in \mathcal{P}_j} \!\!
    x_{j,k}^{i, (t - d_{ij}^{\text{LG},(t)})}\!\!,\,\,
    w_{j,k}^{(t+1)} \!\!\!\leftarrow \!\!\frac{1}{|\mathcal{P}_j|} \!\sum_{i \in \mathcal{P}_j}
    u_{j,k}^{i, (t - d_{ij}^{\text{LG},(t)})}
\end{align}
In the MPC setting we further require $d_{\max} < K$, so that a stale plan remains
valid within the prediction horizon.

\subsection{Failure Mechanisms}
\label{sec:failure_mechanisms}
Delay enters at both rounds, and the algorithm is unable to distinguish stale information from current information. Three consequences follow from the updates above. First, the global update weights every local copy equally at $1/|\mathcal{P}_j|$, so a
copy produced $d$ iterations ago pulls $z_{j,k}$ toward a state no agent still holds.
Second, $r_{i,k}^{(t)}$ is formed against a stale $\tilde{z}_{i,k}^{(t)}$ and thus
mixes genuine disagreement with staleness, causing both the convergence check and any
residual-driven adaptation to misread the algorithm's state. Third, the dual variables
integrate this contaminated mismatch, so the error accumulates. All three are properties of the update rules, so no fixed $\rho$ and no residual-driven schedule can mitigate them (refer to Fig.~\ref{fig:experiment_1_delay_trend}).

\begin{table}[t]
\centering
\caption{Notation Summary}
\label{tab:notation_summary}
\renewcommand{\arraystretch}{1.5} 
\scriptsize
\begin{tabular}{c|p{5.8cm}}
\textbf{Symbol} & \textbf{Description} \\
\hline
$N$ & Number of agents \\
$K$ & Time horizon \\
$\mathcal{N}_i$ & Neighbors of agent $i$ \\
$x_{i,k}$, $u_{i,k}$  & State and control of agent $i$ at time $k$ \\
$x_{j,k}^{i}$, $u_{j,k}^{i}$ & Agent $i$'s local copy of $j$'s state and control at $k$ \\
$\tilde{x}_{i,k}$, $\tilde{u}_{i,k}$ & Augmented state and control at $k$: $[x_{j,k}^{i}]_{j \in \mathcal{N}_i \cup \{i\}}$ \\
$z_{j,k}$, $w_{j,k}$ & Global consensus state and control for agent $j$ at $k$ \\
$\tilde{z}_{i,k}$, $\tilde{w}_{i,k}$ & Vectors $[z_{j,k}]_{j \in \mathcal{N}_i \cup \{i\}}$ and $[w_{j,k}]_{j \in \mathcal{N}_i \cup \{i\}}$ \\
$J_i(x_i, u_i)$ & Local cost for agent $i$ \\
$g_i(\tilde{x}_{i,k})$ & Constraints on augmented state at $k$ \\
$y_{i,k}$, $\lambda_{i,k}$ & Dual variables for $\tilde{x}_{i,k} = \tilde{z}_{i,k}$ and $\tilde{u}_{i,k} = \tilde{w}_{i,k}$ \\
$\rho$, $\mu$ & Penalty parameters \\
$r_{i,k}^{(t)}$ & Primal residual at ADMM iter.\ $t$ \\
$\epsilon_{\text{pri}}$ & Convergence threshold for residual \\
$d_{ij}^{LG,(t)}$, $d_{ij}^{GL,(t)}$ & Delay from agent $j$ to $i$ (and vice versa) at iter.\ $t$ \\
$d_{\max}$ & Maximum communication delay 
\end{tabular}
\end{table}

\subsection{Delay-Aware Parameter Adaptation}
To enhance robustness under communication delays, we propose a \textit{Delay-Aware} ADMM variant, where each agent dynamically adjusts its penalty parameters based on the freshness of received data. Specifically, the penalties $\rho_{j,k}^{i}$ and $\mu_{j,k}^{i}$, for how agent $i$'s optimization considers agent $j$'s information at time $t$, are scaled inversely with the number of delay steps since the last received update:
\begin{align}
\rho_{i,j}^{(t)} = \frac{\rho_{base}}{1 + d_{ij}^{\text{LG},(t)}}, \quad
\mu_{i,j}^{(t)} = \frac{\mu_{base}}{1 + d_{ij}^{\text{LG},(t)}}
\end{align}

This ensures that stale local variables contribute less to the optimization, while more recent data is prioritized.

\paragraph{Local Problem.} Each agent $i$ solves the following delay-weighted local subproblem at iteration $t$:
\begin{align}
\min_{\tilde{x}_i, \tilde{u}_i} \quad 
& J_i(x_i, u_i) 
+
\sum_{k \in \mathcal{K}} 
\frac{\rho_{j,k}^{i}}{2} \left\| \tilde{x}_{j,k}^{i} - \tilde{z}_{j,k}^{(t - d_{ij}^{\text{GL},(t)})} + \frac {y_{i,k}}{\rho_{j,k}^{i}} \right\|^2 \nonumber \\
& \qquad +
\frac{\mu_{j,k}^{i}}{2} \left\| \tilde{u}_{j,k}^{i} - \tilde{w}_{j,k}^{(t - d_{ij}^{\text{GL},(t)})} + \frac{\lambda_{i,k}}{{\mu_{j,k}^{i}}}  \right\|^2  \\
\text{s.t.} \quad 
& \text{dynamics, state, input, and inter-agent constraints.} \nonumber
\end{align}

This formulation adjusts the trust placed on each consensus variable according to its freshness using the weights. 

We derive the new update rule for the global consensus variable $z$ under time-varying penalty parameters; an identical derivation applies to the control variable $w$. As in the standard ADMM formulation, the global update is obtained by minimizing the augmented Lagrangian with respect to $z$: 
\begin{align}
    \mathcal{L}(x, z, y) =& \sum_{i = 1}^N f_i(x_i) +\sum_{j \in \mathcal{N}_i}  y_j^i{}^\top (x_j^i - z_j) \\
    & \quad + \sum_{j \in \mathcal{N}_i} \frac{\rho_j^i}{2} \| x_j^i - z_j \|_2^2 
    \nonumber
\end{align}

The above minimization can be decoupled for a specific $z_\ell$, where $\ell = 1,\dots,N$ and further simplified to:
\begin{align}
z_\ell = \argmin \sum_{i = 1}^N \sum_{j \in \mathcal{N}_i : j = \ell} - y_j^i{}^\top z_j + \frac{\rho_j^i}{2} z_j^\top z_j - \rho_j^i x_j^i{}^\top z_j
\end{align}

If we define the set $\mathcal{P}_i = \{j : i \in \mathcal{N}_j\}$ (the agents $j$ for which agent $i$ belongs to their neighborhoods), the above minimization reduces to: 
\begin{equation}
z_\ell = \argmin \sum_{i \in \mathcal{P}_\ell} - y_\ell^i{}^\top z_\ell + \frac{\rho_\ell^i}{2} z_\ell^\top z_\ell - \rho_\ell^i x_\ell^i{}^\top z_\ell
\end{equation}

We can obtain the minimum of the above unconstrained convex optimization expression by setting the gradient to zero, which yields the following: 

\begin{equation}
z_\ell^{(k+1)} = \frac{\sum_{i \in \mathcal{P}_\ell} y_\ell^i{}^{(k)} + \rho_\ell^i x_\ell^i{}^{(k+1)}}{\sum_{i \in \mathcal{P}_\ell} \rho_\ell^i}
\end{equation}
The dual variable $y_\ell^i{}^{(k)}$ can be further simplified based on the defined dual update as the following:
\begin{align}
\sum_{i \in \mathcal{P}_\ell} y_\ell^i{}^{(k+1)} 
&= \sum_{i \in \mathcal{P}_\ell} \left( y_\ell^i{}^{(k)} + \rho_\ell^i x_\ell^i{}^{(k+1)} \right) - \sum_{i \in \mathcal{P}_\ell} \rho_\ell^i z_\ell^{(k+1)} \nonumber 
\\
=& \sum_{i \in \mathcal{P}_\ell} y_\ell^i{}^{(k)} + \rho_\ell^i x_\ell^i{}^{(k+1)} \nonumber 
\\ 
& \quad \quad \quad - 
\rho_\ell^i \frac{\sum_{m \in \mathcal{P}_\ell} \left( y_\ell^m{}^{(k)} + \rho_\ell^m x_\ell^m{}^{(k+1)} \right)}{\sum_{m \in \mathcal{P}_\ell} \rho_\ell^m} \nonumber \\
=& \sum_{i \in \mathcal{P}_\ell} y_\ell^i{}^{(k)} + \sum_{i \in \mathcal{P}_\ell} \rho_\ell^i x_\ell^i{}^{(k+1)} 
\\
&- \! \sum_{i \in \mathcal{P}_\ell} \rho_\ell^i \cdot \frac{\sum_{m \in \mathcal{P}_\ell} \left( y_\ell^m{}^{(k)} + \rho_\ell^m x_\ell^m{}^{(k+1)} \right)}{\sum_{m \in \mathcal{P}_\ell} \rho_\ell^m} = 0\nonumber 
\end{align}

As a result, the new delayed-aware global updates are simplified to the following weighted averaging rules for both the state and control variables:
\begin{equation}
z_\ell^{(k+1)} = \frac{\sum_{i \in \mathcal{P}_\ell} \rho_\ell^i x_\ell^i{}^{(k+1)}}{\sum_{i \in \mathcal{P}_\ell} \rho_\ell^i}, 
w_\ell^{(k+1)} = \frac{\sum_{i \in \mathcal{P}_\ell} \mu_\ell^i u_\ell^i{}^{(k+1)}}{\sum_{i \in \mathcal{P}_\ell} \mu_\ell^i}
\end{equation}

\paragraph{Dual Update} Dual variables are updated using the adaptive penalties:
\begin{align}
y_{j,k}^{i,(t+1)} &= y_{j,k}^{i, (t)} + \rho_{j,k}^{i} \left( x_{j,k}^{i,(t+1)} - z_{j,k}^{(t+1)} \right) \\
\lambda_{j,k}^{i,(t+1)} &= \lambda_{j,k}^{i (t)} + \mu_{j,k}^{i} \left( u_{j,k}^{i,(t+1)} - w_{j,k}^{(t+1)} \right)
\end{align}

This adaptive penalty serves as \textit{delay-aware regularization}: the optimizer upweights recent messages and softly downweights delayed data, often misaligned due to staleness, thereby limiting its influence on updates and improving convergence stability under unreliable communication.

In contrast to prior asynchronous ADMM variants that rely on inflated safety constraints \cite{ferranti2022distributed}, partial neighborhood information \cite{guo2017impact}, delay-agnostic heuristics such as residual balancing \cite{wohlberg2017admm}, or parameter adaptation schemes based on local sharpness of the objective \cite{xu2017admm}, our method dynamically adjusts local subproblem penalty parameters and defines a new global consensus update based on real-time delay statistics. This enables agents to down-weight outdated information in a principled and distributed manner, without requiring offline retraining or predefined delay bounds.

 \emph{Remark (Convergence and optimality):} We emphasize that
DA-ADMM carries no optimality or convergence guarantees.
Such guarantees are fundamentally hard to obtain here since the multi-robot control problem~\eqref{eq:initial_multi_agent_problem} is non-convex due to collision avoidance and nonlinear dynamics. In such settings, non-convex ADMM converges only under structural assumptions that are not present in our setup \cite{wang2019global, ferranti2022distributed, sun2023two,   saravanos2026distributedcs}, while analyses of asynchronous ADMM are largely restricted to convex
problems~\cite{zhang2014asynchronous, chang2016asynchronous}.
Our time-varying, delay-driven penalties further place the method outside standard analysis frameworks. Nevertheless, non-convex consensus ADMM has repeatedly proven reliable in large-scale multi-robot problems~\cite{saravanos2023distributed_ddp, ferranti2022distributed, saravanos2024dmpcs}.
The design of DA-ADMM is therefore driven primarily by how
delay corrupts the consensus, residual, and dual updates
(Sec.~\ref{sec:failure_mechanisms}), with its effectiveness in this domain
established empirically in Sec.~\ref{sec:experiments}.

\section{Simulation Experiments}
\label{sec:experiments}

\begin{table}[bt]
\centering
\caption{Baseline Comparison}
\label{tab:baseline_comparison}
\renewcommand{\arraystretch}{1.1}
\setlength{\tabcolsep}{4pt}  
\small                      
\begin{tabular}{l|c|c|c|c}
\textbf{Method} & \textbf{Coord.} & \textbf{Delay} & \textbf{Adaptive} & \textbf{Comm.} \\
\hline
DA-ADMM (Ours)            & \cmark & \cmark & \cmark & Med \\
\hline
LB-ADMM (Ours)         & \cmark & \cmark & \xmark & Med \\
\hline
RB-ADMM \cite{wohlberg2017admm}   & \cmark & \xmark & \cmark & Med \\
\hline
FP-ADMM \cite{saravanos2024dmpcs}  & \cmark & \xmark & \xmark & Med \\
\hline
FC-Optimization \cite{luis2020online}            & \xmark & \xmark & \xmark & Low
\end{tabular}
\end{table}

To evaluate the effectiveness of DA-ADMM, we conduct extensive simulation evaluations comparing it against the following established baselines (Table~\ref{tab:baseline_comparison}):

\textit{Lower-Bound (LB) ADMM.} Penalty parameters are initialized according to a delay-aware adaptation rule, $\rho = \rho_{\text{base}} / (1 + d_{\text{max}})$, with $d_{\text{max}}$ denoting the maximum number of consecutive delay steps. These parameters remain fixed throughout the trial, offering some robustness to worst-case delays but lacking online adaptability.

    \textit{Residual Balancing (RB) ADMM.} An adaptive scheme in which penalty parameters are updated based on the ratio between primal and dual residuals, as proposed in \cite{wohlberg2017admm}, and does not incorporate any delay statistics.

    \textit{Fixed Parameter (FP) ADMM.} A constant set of penalty parameters is used across all iterations, independent of delay-related information \cite{saravanos2024dmpcs}.

    \textit{Fixed-Constraint (FC) Optimization.} Each agent plans its trajectory independently by treating the predicted trajectories of its neighbors as fixed constraints, with no consensus mechanism \cite{luis2020online}. Constraints are added selectively based on predicted interactions to reduce conservatism.

All tests are run in Python on a MacBook Pro with an M4 Pro CPU and 48~GB of memory. The optimization problems are formulated using \texttt{CVXPY}~\cite{diamond2016cvxpy} and solved with the \texttt{OSQP} solver~\cite{stellato2020osqp} with warm-starting enabled. The solver's internal iterations is capped at 100{,}000, $dt$ is set to 0.075 and the prediction horizon $K = 40$. A trial is declared infeasible if no solution is found within the specified tolerance before reaching this limit. After optimization, each trajectory is verified for collisions to ensure feasibility and safety. When an agent is delayed, it's latest communicated solution will be used by other agents in the neighborhood until it becomes available again. Finally, agents are not allowed to have more than $d_{max}$ communication delay steps.

 \subsection{2D Double Integrator Circle Formation (Traj-Opt)}

This example evaluates the impact of varying delay patterns on coordination performance in a trajectory optimization setting. The scenario involves robots modeled as linear double integrators arranged in a circular formation, where each robot’s goal state corresponds to the initial state of the robot directly opposite it. The setup requires tight coordination to satisfy challenging collision avoidance constraints in a single-shot optimization. We perform 20 trials with base penalty parameters set to $\rho_x = 0.1$ and $\rho_u = 0.001$ respectively. The optimization is performed using ADMM over 30 outer iterations, with each iteration containing 5 inner SQP steps to linearize the collision avoidance constraints. A trial is deemed successful if all robots reach their respective goals collision-free and within a tolerance of 0.1.
Due to the stringent coordination demands of this scenario, the fixed-constraint distributed optimization baseline consistently fails to make progress and results in deadlock across all tested delay conditions. As a result, it is excluded from the comparisons in this experiment.

\textit{Qualitative Results:} Fig.~\ref{fig:experiment_1_trajopt} illustrates representative trajectories for different methods. Under a 40\% communication delay probability, DA-ADMM (Fig.~\ref{fig:experiment_1_trajopt}b) successfully converges to a solution that closely matches that of FP-ADMM under perfect communication (Fig.~\ref{fig:experiment_1_trajopt}a). In contrast, both FP-ADMM and LB-ADMM fail to converge under the same delay conditions and iteration budget. 

\begin{figure*}[t]
    \centering
    \includegraphics[width=\linewidth]{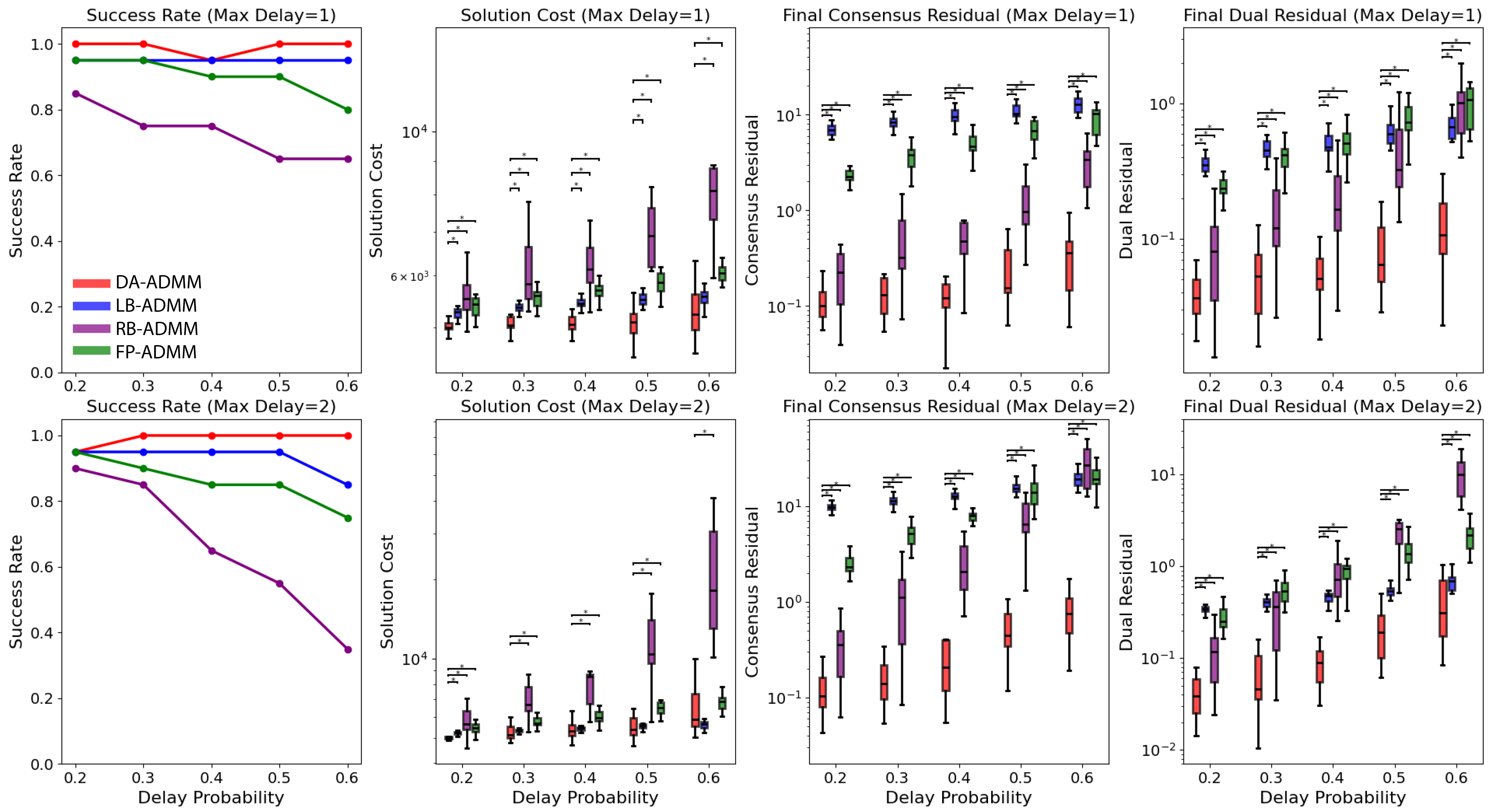}
    \caption{Effect of different delay patterns on distributed optimization approaches in the 2D \textbf{Double Integrator Experiment}. Top and bottom rows correspond to maximum consecutive delay steps of 1 and 2. The columns correspond to success rate (higher is better), total solution cost (lower is better), final consensus residual (lower is better), and final dual residual (lower is better). The asterisks ($*$) represent statistical significance between DA-ADMM and the baselines with a $p < 0.05$. DA-ADMM provides higher success rate and significantly better solution quality compared to other baselines across both delay patterns. This improvement is largely due to significantly lower final primal and dual residuals.}
    \label{fig:experiment_1_delay_trend}
\end{figure*}

\textit{Quantitative Results:} 
Fig.~\ref{fig:experiment_1_delay_trend}(a) and (b) show that success rate and solution quality deteriorate as
delay probability and maximum delay steps increase, consistent with the mechanisms of
Sec.~III-B. DA-ADMM maintains the highest success rates across all delay levels,
followed by LB-ADMM, while FP-ADMM and RB-ADMM decline sharply regardless of the delay
bound, with correspondingly elevated primal and dual residuals in
Figs.~\ref{fig:experiment_1_delay_trend}(c) and (d). Differences of a single trial are within sampling variation at $n=20$.

Notably, RB-ADMM achieves the second lowest residuals yet markedly worse solution quality. Its adaptation is driven solely by the primal-to-dual residual ratio, precisely
the quantity contaminated by staleness. Reading inflated primal residuals as poor convergence, it relaxes the consensus penalty, so residuals fall because agreement is
enforced loosely rather than because the agents agree. RB-ADMM is thus the clearest evidence for the second failure mechanism: adaptation is not the missing ingredient, since a method that adapts on a corrupted signal underperforms a fixed penalty. DA-ADMM instead weights by information age in both the penalty and the averaging rule,
compensating for stale updates while preserving consensus with timely agents.

\begin{table*}[t]
\centering
\caption{Performance under different delay probabilities in the \textbf{Dubins Car Circle Formation Experiment}.
Quantities in parentheses denote \% change vs the same method with perfect communication ($P_{\text{delay}}{=}0$). *Note that in the zero delay case, all ADMM baselines except for RB-ADMM become identical.}
\label{tab:mpc_circle_results}
\setlength{\tabcolsep}{3pt}
\scriptsize
\begin{tabular}{@{}lcccc|cccc|cccc@{}}
& \multicolumn{4}{c|}{$P_{\text{delay}}{=}0^*$} & \multicolumn{4}{c|}{$P_{\text{delay}}{=}0.2$} & \multicolumn{4}{c}{$P_{\text{delay}}{=}0.6$} \\
\cmidrule(lr){2-5}\cmidrule(lr){6-9}\cmidrule(lr){10-13}
\textbf{Algorithm}
& \shortstack{Success\\rate}
& \shortstack{Makespan\\(sec)}
& \shortstack{Primal\\residual}
& \shortstack{Dual\\residual}
& \shortstack{Success\\rate}
& \shortstack{Makespan\\(sec)}
& \shortstack{Primal\\residual}
& \shortstack{Dual\\residual}
& \shortstack{Success\\rate}
& \shortstack{Makespan\\(sec)}
& \shortstack{Primal\\residual}
& \shortstack{Dual\\residual} \\
\midrule
\textbf{DA-ADMM}
& \textbf{10/10}
& \textbf{6.32}
& \textbf{$1.89\!\times\!10^{-14}$}
& \textbf{1.33}
& \textbf{10/10}
& \textbf{6.14} \pgain{$-2.89\%$}
& \textbf{6.23}
& 1.68
& \textbf{9/10}
& \textbf{6.46} \ploss{$+2.19\%$}
& \textbf{24.00}
& 2.34 \\

\textbf{LB-ADMM}
& \textbf{10/10}
& \textbf{6.32}
& $1.25\!\times\!10^{-14}$
& \textbf{1.33}
& \textbf{10/10}
& 6.58 \ploss{$+4.03\%$}
& 8.67
& \textbf{0.55}
& 6/10
& 6.49 \ploss{$+2.65\%$}
& 33.80
& \textbf{0.81} \\

\textbf{FP-ADMM}
& \textbf{10/10}
& \textbf{6.32}
& $1.89\!\times\!10^{-14}$
& \textbf{1.33}
& 9/10
& 6.86 \ploss{$+11.07\%$}
& 10.20
& 1.71
& 4/10
& 6.97 \ploss{$+9.78\%$}
& 45.40
& 3.04 \\

\textbf{RB-ADMM}
& \textbf{10/10}
& 7.28
& $1.25\!\times\!10^{-14}$
& 1.34
& 8/10
& 6.50 \pgain{$-11.03\%$}
& 7.58
& 2.64
& 0/10
& -
& -
& - \\

\textbf{FC-Opt}
& \textbf{10/10}
& 8.76
& -
& -
& 6/10
& 10.38 \ploss{$+16.9\%$}
& -
& -
& 0/10
& -
& -
& - 
\end{tabular}
\vspace{-2em}
\end{table*}
\textit{Value of Communication:} The value of inter-agent communication lies not in its
frequency or recency alone, but in the optimizer's ability to reason over the age of
what it receives. With delay-aware weighting, high-quality solutions are achievable even
when most iterations rely on stale information: DA-ADMM sustains a 100\% success rate in
this experiment at 60\% delay probability. The contrast with LB-ADMM isolates what
matters. LB-ADMM is also delay-aware, but commits to a single conservative penalty set
by $d_{\max}$, and recovers only part of the gap
(Fig.~\ref{fig:experiment_1_delay_trend}(b)). Robustness therefore comes not from
anticipating the worst case, but from adapting to the delay observed at each iteration.

 \subsection{2D Dubins Car Circle Formation (MPC)}
In this experiment, we evaluate the proposed DA-ADMM algorithm in a receding-horizon Model Predictive Controller (MPC) that incorporates online feedback from observations. At each planning cycle, the MPC runs 10 ADMM iterations, each comprising a single SQP inner iteration. Inter-agent communication is subject to delays, consistent with the previous experiment.

To reduce the computation, each robot’s neighborhood is dynamically defined at every planning step as its $N_{\text{neigh}} = 3$ nearest neighbors. This allows agents to reason about the most relevant interactions while keeping the problem size tractable. The robots are modeled using Dubins car dynamics, which are linearized around the solution from the previous time step. In rare instances where the optimization becomes infeasible, the robot executes a fallback safety maneuver with maximum linear deceleration and zero angular velocity. This safety policy remains in effect until a feasible plan is recovered, after which normal control resumes.

\begin{figure*}[t]
    \centering
    \includegraphics[width=.93\linewidth]{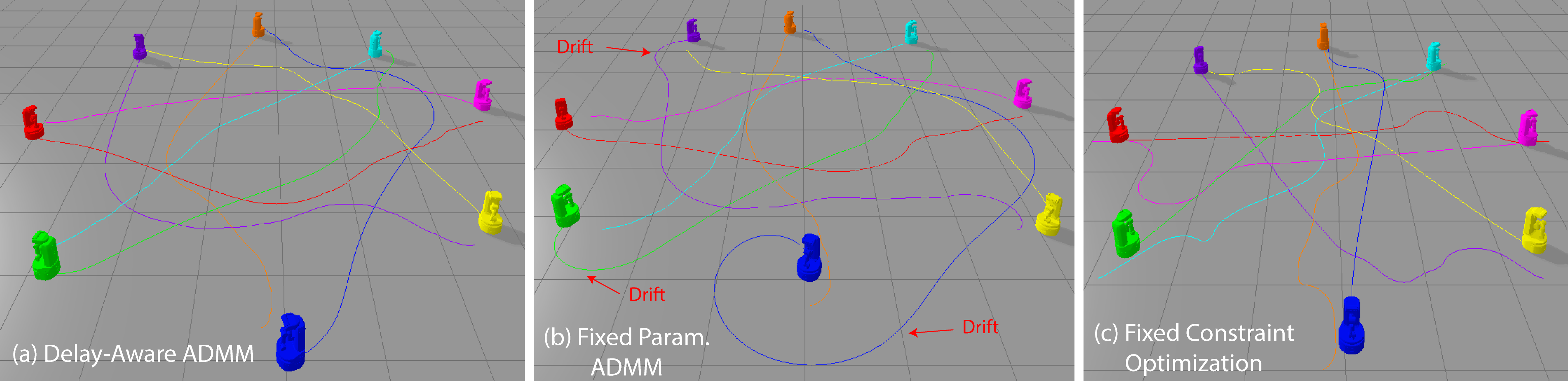}
    \caption{Visual demonstration of the \textbf{Dubins Car Circle Formation Experiment} in PyBullet. Note that DA-ADMM (a) achieves significantly better solution quality relative to FP-ADMM (b) and FC-Opt (c). FP-ADMM suffers from solution drift under delay as shown in the green, blue, and purple agent trajectories in (b).}
    \label{fig:mpc_circle}
    \vspace{-1em}
\end{figure*}

As summarized in Table~\ref{tab:mpc_circle_results}, DA-ADMM consistently outperforms other baselines in terms of success rate and task completion time. While the MPC formulation improves the performance of FP-ADMM and FC-Opt, it does not eliminate the performance gap relative to DA-ADMM. As shown in Fig.~\ref{fig:mpc_circle}, the solution of FP-ADMM drifts during the execution relative to the proposed approach. This is primarily due to poor ADMM residual convergence at each planning step, which leads to compounding deviations or collision failures. Also note the poorer performance of RB-ADMM under delays compared with other ADMM baselines. Because RB-ADMM overreacts to residuals contaminated by stale information, it induces oscillatory updates in $\rho$ and yields worse convergence than a well-chosen fixed-$\rho$ scheme \cite{xu2017adaptive}.

\textit{Robustness Under Delays in ADMM vs FC-Opt:} Another important insight is the comparison between the Fixed Constraint Optimization (FC-Opt) approach and the ADMM approaches. Notably, the success rate of FC-Opt deteriorates significantly as communication delay increases, whereas the ADMM variants maintain stronger performance. In the FC-Opt approach, each agent computes its solution based on the current information available from its neighbors. When delays occur, outdated neighbor information is used, leading to degraded coordination over time. In contrast, ADMM iteratively refines consensus among agents, allowing delayed information to still influence subsequent iterations. This iterative structure provides greater robustness to delays, but comes at the cost of increased communication overhead.

\subsection{2D Dubins Car Warehouse Operation (MPC)}
\begin{figure}[t]
    \centering
    \includegraphics[width=.9\linewidth]{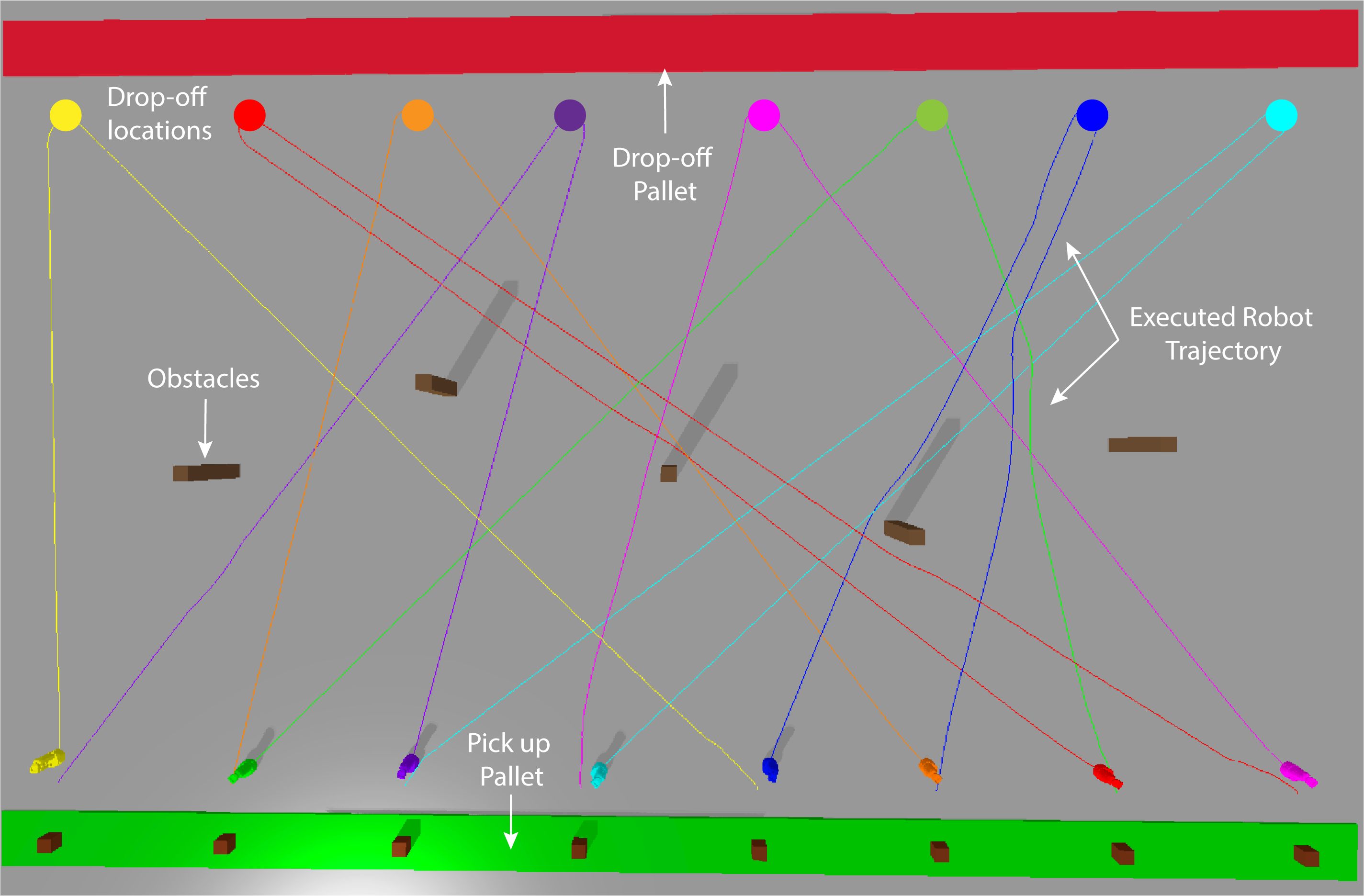}
    \caption{DA-ADMM results for the \textbf{Dubins Car Warehouse Experiment} with continuous task assignment.}
    \label{fig:warehouse}
\end{figure}
This experiment simulates a more realistic warehouse environment, where robots are tasked with continuously transporting boxes between two opposing conveyor belts (Fig.~\ref{fig:warehouse}). As soon as a robot reaches its goal state, it is immediately given a new target location at the opposite conveyor, creating an ongoing cycle of interactions and planning updates. A key feature of this setup is the asynchronous and continuous task assignment, in contrast to many prior works that focus solely on single point-to-point navigation. This dynamic tasking mechanism better reflects real-world scenarios, where the number of robot-robot interactions can vary over time. As summarized in Table \ref{tab:mpc_warehouse_results}, DA-ADMM retains its advantage relative to other baselines under this realistic setting, where interactions vary over time.  

\begin{table}
\centering
\caption[Performance under different delays in the Dubins Car Warehouse Experiment]{Performance under different delay probabilities in the \textbf{Dubins Car Warehouse Experiment}.
Quantities in parentheses denote \% change vs the same method with perfect communication ($P_{\text{delay}}{=}0$).}
\label{tab:mpc_warehouse_results}
\setlength{\tabcolsep}{3pt}
\scriptsize
\begin{tabular}{@{}lcccc|cccc@{}}
& \multicolumn{4}{c|}{$P_{\text{delay}}{=}0.3$} 
  & \multicolumn{4}{c}{$P_{\text{delay}}{=}0.6$}  \\
\cmidrule(lr){2-5}\cmidrule(lr){6-9}
\textbf{Algorithm}
& \shortstack{Succ.\\rate}
& \shortstack{Makespan\\(sec)}
& \shortstack{Primal\\res.}
& \shortstack{Dual\\res.}
& \shortstack{Succ.\\rate}
& \shortstack{Makespan\\(sec)}
& \shortstack{Prim.\\res.}
& \shortstack{Dual\\res.} \\
\midrule
\textbf{DA-ADMM}
& \textbf{5/5}
& \textbf{26.95}
&  \textbf{1.48}
&  \textbf{0.08}
& \textbf{4/5}
& \textbf{26.74}
& \textbf{4.14}
& \textbf{0.23}
 \\
\textbf{FP-ADMM}
& \textbf{5/5}
& 25.96
& 9.59
& 1.81
& 3/5
& 27.65
& 25.7
& 4.85
 \\
\textbf{FC-Opt}
& 0/5
& -
& -
& -
& 0/5
& -
& -
& -
\end{tabular}
\end{table}

\subsection{3D Drone Model Demo (MPC)}
\begin{table*}[t]
\centering
\caption{Performance under different delay probabilities in the \textbf{3D Drone Experiment}.
Quantities in parentheses denote \% change vs the same method with perfect communication ($P_{\text{delay}}{=}0$).}
\label{tab:mpc_drone_results}
\setlength{\tabcolsep}{3pt}
\scriptsize
\begin{tabular}{@{}lccc|ccc|ccc@{}}
& \multicolumn{3}{c|}{$P_{\text{delay}}{=}0^*$} & \multicolumn{3}{c|}{$P_{\text{delay}}{=}0.3$} & \multicolumn{3}{c@{}}{$P_{\text{delay}}{=}0.6$} \\
\cmidrule(lr){2-4}\cmidrule(lr){5-7}\cmidrule(lr){8-10}
\textbf{Algorithm}
& \shortstack{Success\\rate}
& \shortstack{Makespan\\(sec)}
& \shortstack{Comp. time\\(sec)}
& \shortstack{Success\\rate}
& \shortstack{Makespan\\(sec)}
& \shortstack{Comp. time\\(sec)}
& \shortstack{Success\\rate}
& \shortstack{Makespan\\(sec)}
& \shortstack{Comp. time\\(sec)} \\
\midrule
\textbf{DA-ADMM}
& \textbf{5/5}
& \textbf{6.68}
& \textbf{6.04}
& \textbf{5/5}
& \textbf{6.68} \ploss{$0.0\%$}
& \textbf{6.50} \ploss{$+7.62\%$}
& \textbf{3/5}
& \textbf{8.03} \ploss{$+18.3\%$}
& \textbf{6.82} \ploss{$+12.13\%$} \\
\textbf{LB-ADMM}
& \textbf{5/5}
& \textbf{6.68}
& 6.53
& 4/5
& \textbf{6.68} \ploss{$0.0\%$}
& 6.53 \ploss{$0.0\%$}
& 2/5
& 12.56 \ploss{$+61.2\%$}
& 7.30 \ploss{$+13.9\%$} \\
\textbf{FP-ADMM}
& \textbf{5/5}
& \textbf{6.68}
& 6.11
& 2/5
& \textbf{6.68} \ploss{$0.0\%$}
& 6.53 \ploss{$+6.38\%$}
& 0/5
& -
& - \\
\textbf{FC-Opt}
& 0/5
& -
& -
& 0/5
& -
& -
& 0/5
& -
& - 
\end{tabular}
\vspace{-1em}
\end{table*}

\begin{figure}[t]
    \centering
    \includegraphics[width=\linewidth]{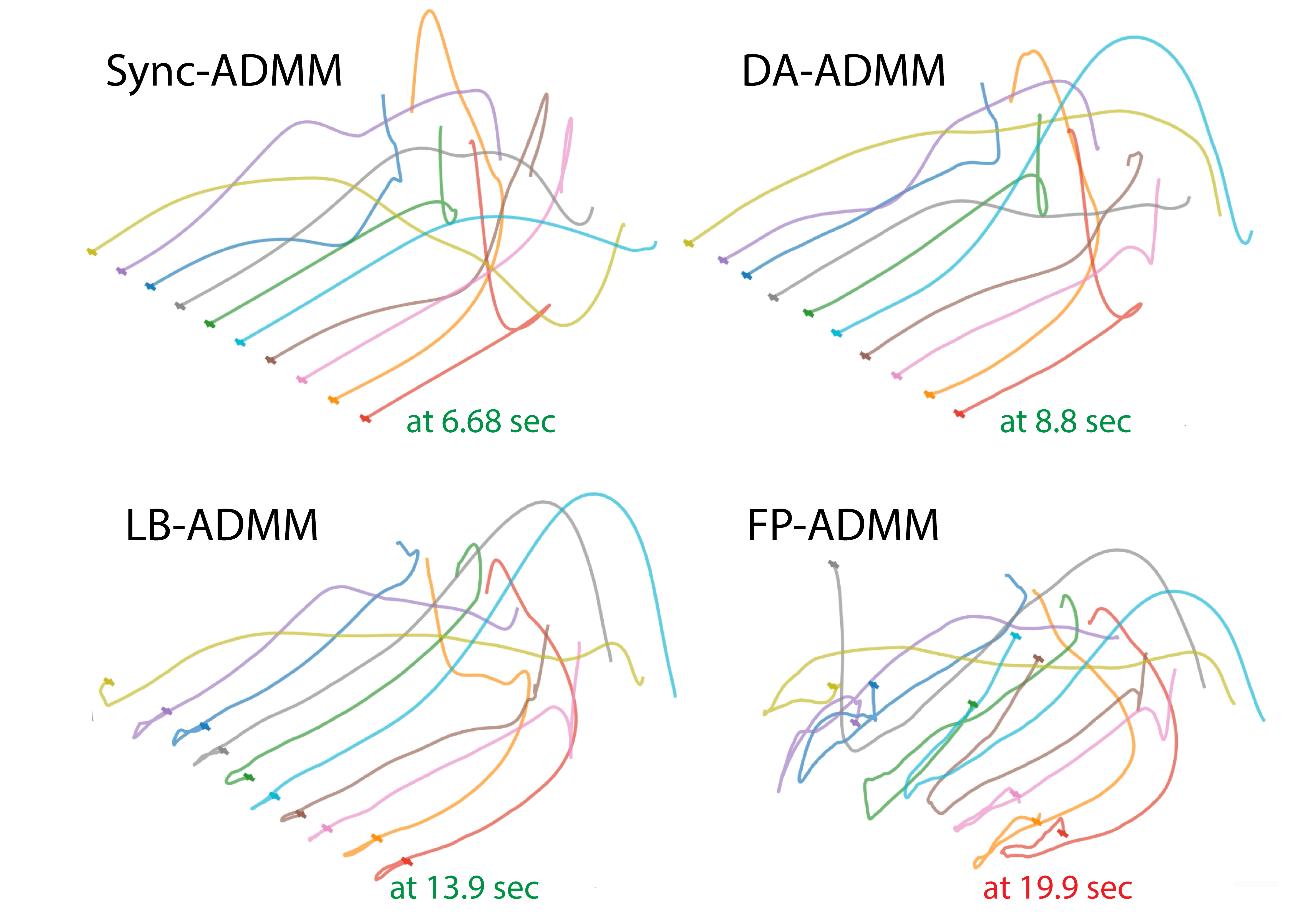}
    \caption[Qualitative results of the 3D drone experiment]{Qualitative results for the \textbf{3D Drone Experiment}. DA-ADMM is the only method that can complete the task with similar performance to the synchronized ADMM with perfect communication. Green denotes successful task completion, and red denotes failure within max allowable time.}
    \label{fig:drone_3d}
\end{figure}

We evaluate our method on a challenging 3D MPC task with $N = 10$ agents, each communicating with $N_{neigh} = 3$ neighbors, and performing 10 ADMM iterations per MPC solve. Each agent uses a linear drone model with 9 states and 6 controls \cite{zhang2025gcbf+}. Under the consensus formulation, the per-agent augmented decision vectors have dimension 36 for states and 24 for controls. As shown in Fig.~\ref{fig:drone_3d}, DA-ADMM successfully completes the task in 8.8 seconds, about 2 seconds slower than synchronized ADMM under ideal communication. In contrast, LB-ADMM takes 13.9 seconds and FP-ADMM fails to converge within the 19.9 seconds time limit, due to compounding consensus error. As summarized in Table~\ref{tab:mpc_drone_results}, DA-ADMM is the most robust to communication delays, while FC-Opt failed in all trials due to infeasibility under tight inter-robot coupling. These results highlight that our delay-aware formulation scales effectively to complex 3D scenarios, extending the benefits observed in 2D to settings where robots are not space-constrained.

\textit{Computational Efficiency:} In our current (unoptimized) prototype, one MPC solve with 10 ADMM iterations takes \(\approx 0.65\,\mathrm{s}\) per robot. With straightforward engineering, parallelizing primal/dual updates across agents, using a low-overhead direct solver interface with warm starts, and running the ADMM consensus loop at a lower rate than the inner MPC controller, we expect the method can be run online~\cite{saravanos2024dmpcs}.

 \section{Conclusions}
This work presents a systematic investigation into the impact of communication delays on distributed multi-robot motion planning and proposes a principled delay-aware adaptation scheme within the consensus ADMM framework. Unlike existing methods that either ignore delay or treat it conservatively, our delay-aware approach dynamically adjusts penalty weights based on delay statistics to selectively trust information during optimization. Our extensive experiments confirm that the delay-aware strategy enables consistently higher success rates, improved convergence, and better solution quality under a range of communication delay patterns. The proposed method maintains feasibility and avoids deadlocks even in tightly coupled scenarios where traditional fixed constraint approaches fail. 

We demonstrate that stale information can still be leveraged when weighted appropriately, challenging the notion that frequent updates are necessary for robust coordination. While this work
attributes degraded information quality to communication delay, delay-aware weighting is complementary to distributed methods addressing other sources of uncertainty, such as stochasticity in the dynamics \cite{saravanos2024dmpcs}, and the two combine
naturally within the same consensus framework.


\bibliographystyle{IEEEtran}
\bibliography{references}
\end{document}